\documentclass{article}
\usepackage{spconf,amsmath,epsfig}
\usepackage[linesnumbered,ruled,vlined]{algorithm2e}
\usepackage{multirow}
\usepackage[hidelinks]{hyperref}
\usepackage{amsfonts}
\usepackage{colortbl}
\definecolor{Gray}{gray}{0.92}
\newcolumntype{g}{>{\columncolor{Gray}}c}
\graphicspath{ {images/} }

\let\OLDthebibliography\thebibliography
\renewcommand\thebibliography[1]{
  \OLDthebibliography{#1}
  \setlength{\parskip}{0pt}
  \setlength{\itemsep}{0pt plus 0.3ex}
}

\begin{document}\sloppy

\def\x{{\mathbf x}}
\def\L{{\cal L}}

\title{Federated Unsupervised Domain Adaptation for Face Recognition}
%
\name{Weiming Zhuang$^{1,3}$, Xin Gan$^{2}$, Xuesen Zhang $^{3}$, Yonggang Wen$^{2}$, Shuai Zhang$^{3}$, Shuai Yi$^{3}$}
\address{$^{1}$S-Lab, NTU, Singapore\, $^{2}$NTU, Singapore\, $^{3}$SenseTime Research \\ weiming001@e.ntu.edu.sg, ygwen@ntu.edu.sg, zhangshuai@sensetime.com}

\maketitle

\begin{abstract}
Given labeled data in a source domain, unsupervised domain adaptation has been widely adopted to generalize models for unlabeled data in a target domain, whose data distributions are different. However, existing works are inapplicable to face recognition under privacy constraints because they require sharing of sensitive face images between domains. To address this problem, we propose federated unsupervised domain adaptation for face recognition, \textit{FedFR}. FedFR jointly optimizes clustering-based domain adaptation and federated learning to elevate performance on the target domain. Specifically, for unlabeled data in the target domain, we enhance a clustering algorithm with distance constrain to improve the quality of predicted pseudo labels. Besides, we propose a new domain constraint loss (\textit{DCL}) to regularize source domain training in federated learning. Extensive experiments on a newly constructed benchmark demonstrate that FedFR outperforms the baseline and classic methods on the target domain by 3\% to 14\% on different evaluation metrics.
\end{abstract}

\section{Introduction}

Despite that face recognition using the deep neural network has achieved outstanding performances \cite{deng2019arcface}, a well-trained model would fail to generalize across different face attributes like age and ethnicity. For example, a model trained with fair-skinned face images in one region would not perform well in another with tan-skinned data \cite{wang2019rfw}. This problem is the \textit{domain shift problem} --- the \textit{source domain} where models are trained share different data distributions with the \textit{target domain} where models are deployed. 


\begin{figure}[h]
    \begin{center}
    \includegraphics[width=\linewidth]{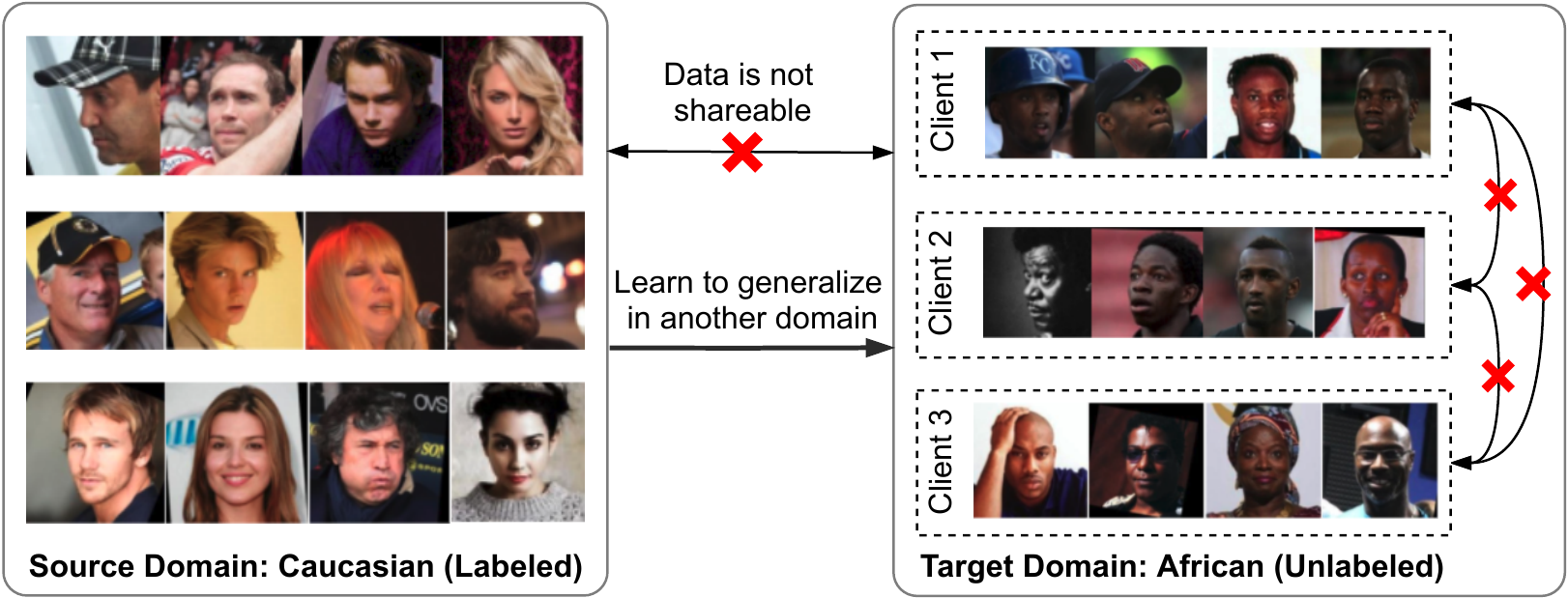}
    \end{center}
    \vspace{-0.5cm}
        \caption{We propose federated unsupervised domain adaptation for face recognition (FedFR) to learn models for the \textit{unlabeled} target domain, under privacy constraints that 1) data are \textit{not shareable} across domains and 2) unlabeled data are collected to edge devices (clients) in the target domain.}
    \label{fig:problem}
    \vspace{-0.3cm}
\end{figure}

Moreover, the domain shift problem is even more challenging in real-world scenarios where data are unlabeled in the target domain and not shareable across domains, as shown in Fig. \ref{fig:problem}. Firstly, data cannot be shared due to increasingly stringent data protection regulations; these regulations have limited data sharing among organizations in countries, especially for sensitive data like face images \cite{gdpr}. Secondly, data are mostly unlabeled in the target domain when they are collected and stored in multiple edge devices. Centralizing them would also imply potential risks of privacy leakage.

Existing methods, however, cannot adequately address these challenges, especially when data is not sharable between domains. An intuitive idea is using only data in the target domain, but collecting and labeling more data is expensive and possibly restricted. Although researchers propose unsupervised domain adaptation methods \cite{sohn2018uda-metric-learning, wang2020uda-clustering-face}, these methods assume data is sharable. To preserve data privacy, Peng et al. \cite{Peng2020fada} leverages federated learning, an emerging distributed training method \cite{fedavg}. However, their method is for image classification and requires multiple source domains to adapt to one target domain, whereas only one source domain is available in our scenario. Image classification have identical classes between domains, whereas face recognition is even more challenging as the identities are different in the two domains.



\begin{figure*}[t]
\begin{center}
\includegraphics[width=\linewidth]{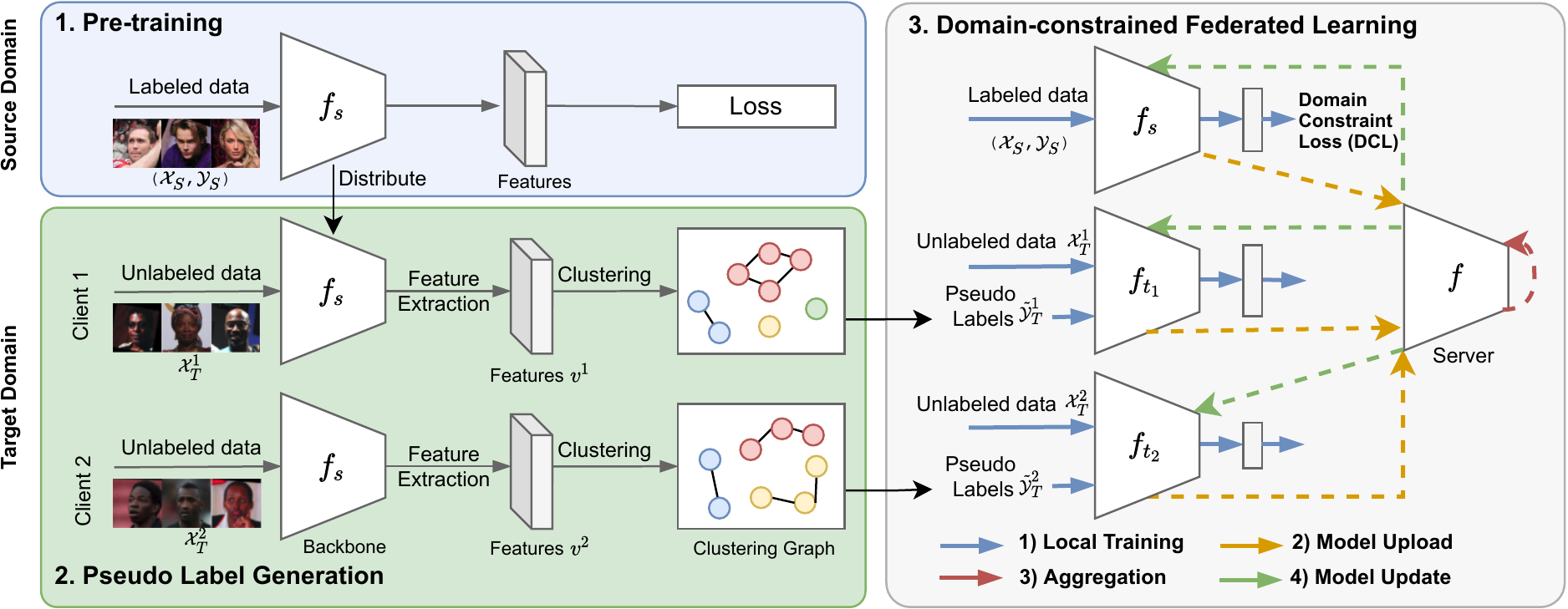}
\end{center}
\vspace{-0.5cm}
   \caption{Overview of our proposed FedFR. It forms an end-to-end training pipeline with three stages: (1) \textit{Pre-training}: train a model $f_s$ in the source domain with data $\mathcal{D}_S=\{\mathcal{X}_S, \mathcal{Y}_S\}$; (2) \textit{Pseudo label generation}: predict pseudo labels $\tilde{\mathcal{Y}}_T$ for target domain data $\mathcal{D}_T=\{\mathcal{X}_T\}$ using clustering; (3) \textit{Domain-constrained Federated learning}: train a model $f$ that generalizes well on the target domain with one client of source domain and two clients (e.g. edge devices) of target domain (for demonstration). We jointly optimize the clustering algorithm and FL to elevate performance on the target domain.}
\label{fig:framework}
\vspace{-0.3cm}
\end{figure*}

In this paper, we propose federated unsupervised domain adaptation for face recognition, \textit{FedFR}, to tackle the domain shift problem under privacy constraints. In particular, we jointly optimize the clustering-based domain adaptation \cite{wang2020uda-clustering-face} and federated learning \cite{fedavg} to achieve compelling performance on the target domain. 
Firstly, we adopt clustering to generate pseudo labels for the unlabeled target domain data. To alleviate false labels caused by confusion of faces in clustering, we optimize a hierarchical clustering algorithm \cite{sarfraz2019finch} by implementing a distance constraint. 
Secondly, we leverage federated learning to iteratively aggregate knowledge learned in source and target domains. In particular, a central server aggregates models that are independently trained in either domain, preserving data privacy by transmitting models instead of raw data. 
Since the source domain could contain much more data than the target domain, we propose domain-constrained federated learning with a new domain constraint loss (DCL) to regularize models trained in the source domain. 
Incorporating these two optimizations, FedFR comprises an end-to-end training pipeline with three stages: (1) source domain pre-training; (2) distance-constrained pseudo label generation; (3) domain-constrained federated learning. 

Besides, we construct a new benchmark for FedFR. Extensive experiments of the benchmark demonstrate the effectiveness of FedFR. It surpasses the baseline and other classic methods by 3\% to 14\% of verification rates and 9\% of identification rank-1 accuracy. We believe that FedFR will shed light on applying federated unsupervised domain adaptation to more computer vision tasks under privacy constraints.

\section{Related Work}


\textbf{Federated Learning} \; Federated learning (FL) is a distributed training technique that learns models with decentralized clients under the coordination of a server by transmitting models instead of raw data to preserve data privacy \cite{fedavg}. Previous studies have applied FL to tasks like person re-identification \cite{zhuang2020fedreid,zhuang2021fedureid}, but the application in face recognition is largely underexplored. Recent studies implement federated face recognition \cite{bai2021federated,niu2021federated}, but they rely on data labels in all clients. Although several studies of FL consider unsupervised domain adaptation \cite{Peng2020fada, song2020uda-federated-privacy}, these methods would fail in face recognition because they are designed for classification tasks. Since the standard FL algorithm, FederatedAveraging (FedAvg) \cite{fedavg}, requires identical models in the server and clients, we exploit Federated Partial Averaging (FedPav) \cite{zhuang2020fedreid} to synchronize only a backbone of the model.





\textbf{Unsupervised Domain Adaptation for Face Recognition} \; Unsupervised domain adaptation (UDA) has received great attention \cite{long2015dan, ganin2016dann}. With the objective of performing well on the unlabeled target domain, existing studies leverage the following UDA methods for face recognition: 
Sohn et al. used domain adversarial discriminator to learn domain-invariant features \cite{sohn2017uda-FR-video}; Luo et al. applied maximum mean discrepancies (MMD) loss to face recognition \cite{luo2018uda-FR}; Wang and Deng proposed a clustering-based method with MMD loss \cite{wang2020uda-clustering-face}. However, these domain alignment approaches require locating data together, violating the privacy constraint of data sharing. In this work, we introduce a new approach, FedFR, to jointly optimize the clustering-based domain adaptation and FL algorithms to elevate performance on the target domain under privacy constraints.


\section{Methodology}
\label{sec:methodology}

This section presents the proposed federated unsupervised domain adaptation for face recognition (FedFR) to address the domain discrepancy between the source and target domains without data sharing.

\subsection{Problem Definition}
\label{sec:problem-definition}

Before illustrating the details of FedFR, we present the problem and the assumptions first. The domain shift problem arises when a model is trained in one domain but deployed to another. These two domains have different data distributions. This paper investigates the domain shift problem under the constraint that data are non-shareable between domains due to privacy protection and data being unbalanced. Specifically, we aim to obtain a model $f$ that delivers high performance on the target domain, given source domain data $\mathcal{D}_S = \{\mathcal{X}_S, \mathcal{Y}_S\}$ and unlabeled target domain data $\mathcal{D}_T = \{\mathcal{X}_T\}$. $\mathcal{D}_S$ and $\mathcal{D}_T$ are under two assumptions: they are located in different places and not shareable; the size of $\mathcal{D}_T$ could be much smaller than $\mathcal{D}_S$. Moreover, the target domain comprises several non-shareable datasets that are collected from multiple edge devices, $\mathcal{D}_T = \{\mathcal{D}_T^1, \cdots , \mathcal{D}_T^k, \cdots , \mathcal{D}_T^K | K \ge 1 \}$, where each client $k$ contains data $\mathcal{D}_T^k$. Thus, we use FedFR to learn a face recognition model $f$ that generalizes on the target domain, with $N=K+1$ decentralized clients, including only one source domain client and $K$ target domain clients.

\subsection{FedFR Overview}

Fig. \ref{fig:framework} provides an overview of FedFR. To obtain a model that delivers high performance on the target domain without cross-domain data sharing, FedFR comprises three stages: source domain pre-training, pseudo label generation, and domain-constrained federated learning. These three stages form an end-to-end training pipeline: (1) We train a face recognition model $f_s$ with the source domain data $\{\mathcal{X}_S, \mathcal{Y}_S\}$. (2) In the target domain, each client $k$ downloads $f_s$, extracts features $v^k$ with local data $\mathcal{X}_T^k$, and clusters $v^k$ to generate pseudo labels $\tilde{\mathcal{Y}}_T^k$. (3) We conduct federated learning with a server to coordinate multiple clients --- the source domain as a client with $(\mathcal{X}_S, \mathcal{Y}_S)$ and all clients in the target domain with $(\mathcal{X}_T^k, \tilde{\mathcal{Y}}_T^k)$ --- to obtain a global model $f$ that generalizes across different domains. 
Next, we present the joint optimization of clustering and federated learning.


\subsection{Pseudo Label Generation}
\label{sec:clustering}

For unlabeled target domain data, we predict their pseudo labels using the pre-trained source domain model and clustering algorithms. As illustrated in the second stage in Fig. \ref{fig:framework}, we first extract features $v^k$ from unlabeled data $\mathcal{D}_T^k$ in each client $k$, using the pre-trained model $f_s$ from the source domain. Then, we apply clustering algorithms on $v^k$ to form clustering graphs. Face images in the \textit{same cluster} are considered to be the \textit{same identity} and are assigned the \textit{same pseudo label}.

The quality of pseudo labels depends on the clustering algorithm, so we propose an enhanced distance-constrained hierarchical clustering algorithm to elevate the performance of clustering to alleviate the confusion of faces. Built on FINCH \cite{sarfraz2019finch}, we implement a new distance regularization to decide whether to merge two clusters, termed Conditional FINCH (C-FINCH).
Regarding each extracted feature as a cluster at the start, C-FINCH merges two clusters if they are first neighbors and their distance is smaller than a threshold $d$. We formulate it as followed:
\begin{equation}
    C(i, j)
    \begin{cases}
        1 & \quad \text{if } N^1(i, j) \text{ and } D_{distance}(i, j) < d \\
        0 & \quad \text{otherwise},
    \end{cases}
    \label{eq:c-finch}
\end{equation}
where $N^1(i, j)$ means that cluster $i$ and $j$ are first neighbors and $D_{distance}(i, j)$ measures the distance between their centroids. Two clusters are first neighbors if their centroids are at the minimum distance or they share the same closest cluster. The distances measure the similarity between two clusters (identities) --- larger distance indicates that two identities are more divergent. We optimize the clustering algorithm by enforcing the distance of first neighbors to be smaller than a threshold $d$ to mitigate the negative impact of incorrectly labeled faces. Finally, we annotate each cluster with a unique pseudo label for the next stage.

\subsection{Domain-constrained Federated Learning}
\label{sec:federated-learning}

To reduce the domain discrepancy without sharing data between domains, we aggregate the knowledge from both domains via federated learning (FL). Regarding the source domain and each edge of the target domain as clients, they perform training collaboratively under the coordination of a central server. As the source domain normally contains much more data than the target one, we propose \textit{domain-constrained federated learning}, to regularize the source domain training with a new \textit{domain constraint loss} (DCL). 

The third stage in Fig. \ref{fig:framework} presents the training flow of domain-constrained FL. The training procedure begins with initializing all clients in both domains with model parameters $\theta^s$ from the source domain pre-trained model $f_s$. It conducts iterative training, where each training round $t$ includes the following steps: (1) \textit{Local Training}: each client conducts $E$ local iterations of training with its local dataset. The source domain client trains with $(\mathcal{X}_S, \mathcal{Y}_S)$ using DCL, while each target domain client $k$ trains with $(\mathcal{X}_T^k, \tilde{\mathcal{Y}}_T^k)$. (2) \textit{Model Upload}: each client transfers the training updates $\theta_{r+1}^k$ to the server. (3) \textit{Aggregation}: the server aggregates these model updates to obtain a new global model with $\theta_{r+1} = \frac{1}{N} \sum_{k=1}^N \theta_{r+1}^k$ . (4) \textit{Model Update}: each client downloads the global model $\theta_{r+1}$ to update its local model for the next round of training. 
  

\textbf{Domain Constraint Loss} To tackle the unbalanced data volume of two domains, we propose DCL to regularize the source domain training. Before delving into the details of DCL, we first analyze the limitations of conventional FL \cite{fedavg}.


FedAvg \cite{fedavg} aims to obtain a global model $f$ with: 
\begin{equation}
    \min_{\theta \in \mathbb{R}^d} f(\theta) :=  \frac{1}{N} \sum_{k=1}^N F_k(\theta),
\label{eq:fl}
\end{equation}
where $N$ is the number of clients and $F_k: \mathbb{R}^d \rightarrow \mathbb{R}$ represents the expected loss over data distribution $\mathcal{D}_k$ of each client $k$: $F_k(\theta) := \mathbb{E}_{x_k \sim \mathcal{D}_k}[f_k(\theta;x_k)]$,
where $x_k$ is the data in client $k$ and $f_k(\theta;x_k)$ is a loss function to train model $\theta$.


FedAvg (Equation \ref{eq:fl}) aggregates model updates with averaging, which equally weighs the importance of both domains. 
However, it is not optimal to deliver high performance on the target domain, especially when the target domain contains \textit{far less} data than the source domain. As a result, we need to regularize the source domain and reinforce the importance of the target domain.
We propose DCL for the source domain training. The loss functions for target and source domains are formulated as followed:
\begin{align}
    \mathcal{L}_{target} &= \mathcal{L}_{face}, \label{eq:target-loss} \\
    \mathcal{L}_{source} &= \mathcal{L}_{face} + \underbrace{\frac{\lambda}{2} \left\Vert \theta^s_r - \theta_{r-1} \right\Vert^2}_{\mathcal{L}_{DCL}},
    \label{eq:source-loss}
\end{align}
where $\mathcal{L}_{face}$ is the loss for face recognition and $\mathcal{L}_{DCL} = \frac{\lambda}{2} \left\Vert \theta^s_r - \theta_{r-1} \right\Vert^2$ denotes domain constraint loss; $\theta^s_r$ is the source domain model to be optimized in round $r$, $\theta_{r-1}$ is the parameters of the global model from the previous round, and $\lambda$ is a control parameter for the intensity of the regularization.

\begin{figure}[t]
\begin{center}
\includegraphics[width=0.9\linewidth]{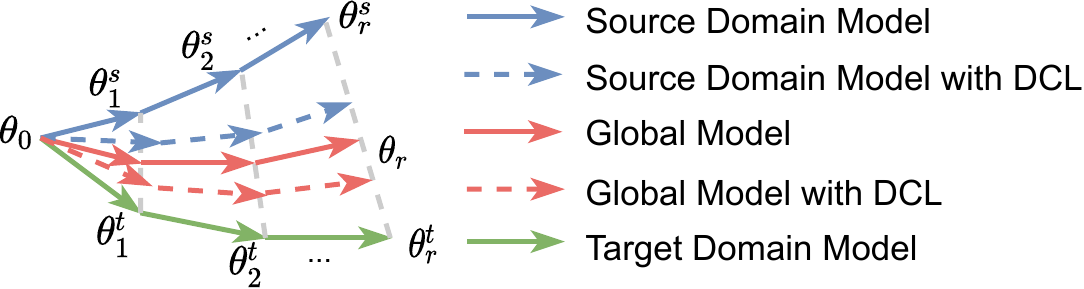}
\end{center}
\vspace{-0.5cm}
  \caption{Illustration of domain constrained loss (DCL). DCL addresses unbalanced data volume between domains by regularizing the source domain model such that the aggregated global model (with DCL) can lean towards the target domain.}
\label{fig:intuition}
\vspace{-0.3cm}
\end{figure}

We depict the intuition of DCL in Fig. \ref{fig:intuition}. In each round, the global model is the averaging of source and target models. We add DCL in the source domain to ensure that the source domain model does not deviate significantly from the global model during training. This is especially relevant when the source domain contains much more data. With DCL in the source domain, the global model (with DCL) can lean towards the target domain, resulting in better performance.

\begin{table*}[ht]
\caption{Statistics of training and evaluation datasets in the newly constructed benchmark. We use the test set to evaluate face verification and construct queries and galleries to evaluate face identification.}
\label{tab:evaluation-data}
\begin{center}
\begin{tabular}{llcccccccccccc}
\hline
\multicolumn{1}{l}{\multirow{2}{*}{Dataset}} &
\multicolumn{1}{l}{\multirow{2}{*}{Domain}} &
\multicolumn{1}{c}{} &
\multicolumn{2}{c}{Train} &
\multicolumn{1}{c}{} &
\multicolumn{2}{c}{Test} &
\multicolumn{1}{c}{} &
\multicolumn{2}{c}{Query} &
\multicolumn{1}{c}{} &
\multicolumn{2}{c}{Gallery}
\\ 
\cline{4-5} \cline{7-8} \cline{10-11} \cline{13-14}
& & & \# IDs & \# Images & & \# IDs & \# Images & & \# IDs & \# Images & & \# IDs & \# Images \\ 
\hline 
Caucasian & Source & & 87,072 & 4,434,177 & & 2,959 & 10,196 & & 2,793 & 2,793 & & 2,958 & 6,387 \\ 
African & Target & & 7,000 & 324,376 & & 2,995 & 10,415 & & 2,865 & 2,865 & & 2,995 & 6,770 \\ 
\hline
\vspace{-1cm}
\end{tabular}
\end{center}
\end{table*}


\begin{table*}[t]
  \caption{Comparison of face verification and identification (Id.) performances on the benchmark. FedFR significantly improves the performance on the target domain (African), even outperforming the Merge model.}
    
  \label{tab:fedfr-l}  
    \begin{center}
    \begin{tabular}{lggggggcccccccc}
    \hline
    \multicolumn{1}{l}{\multirow{3}{*}{Model}} &
    \multicolumn{6}{g}{African (\%)} &
    \multicolumn{1}{c}{} &
    \multicolumn{6}{c}{Caucasian (\%)} 
    \\ 
      \cline{2-7}  \cline{9-14}
    \multicolumn{1}{c}{} &
    
    \multicolumn{4}{g}{Verification} &
    \multicolumn{1}{g}{} &
    \multicolumn{1}{g}{Id.} &
    
    \multicolumn{1}{c}{} &
    
    \multicolumn{4}{c}{Verification} &
    \multicolumn{1}{c}{} &
    \multicolumn{1}{c}{Id.}
    \\ 
    \cline{2-5} \cline{7-7} \cline{9-12} \cline{14-14} 
    \multicolumn{1}{c}{} & Acc & 0.1 & 0.01 & 0.001 & & Rank-1 & & Acc & 0.1 & 0.01 & 0.001 & & Rank-1 \\
    \hline
    Source-Only & 86.65 & 83.27 & 55.93 & 37.80 & & 82.13 & & 94.47 & 95.67 & \textbf{87.53} & \textbf{76.67} & & 93.52 \\ 
    Target-Only & 83.35 & 75.6 & 41.80 & 22.00 & & 73.54 & & 86.23 & 81.93 & 54.83 & 38.27 & & 46.26 \\
    Merge & 88.60 & 87.00 & 66.90 & 48.10 & & 89.04 & & \textbf{94.78} & \textbf{96.60} & 85.63 & 74.43 & & 94.38 \\ 
    Fine-tune & 88.72 & 87.10 & 62.23 & 35.95 & & 84.05 & & 87.73 & 85.87 & 62.87 & 39.77 & & 77.77 \\ 
    DAN & 86.98 & 83.87 & 61.73 & 41.43 & & 83.18 & & 94.07 & 95.43 & 85.20 & 72.13 & & 93.30 \\ 
    DANN & 86.77 & 83.23 & 60.80 & 44.90 & & 82.23 & & 93.98 & 96.03 & 86.23 & 73.97 & & 93.59  \\ 
    \hline
    
    FedFR (N=5) & \textbf{90.53} & \textbf{90.43} & \textbf{73.37} & \textbf{52.33} & & \textbf{90.96} & & 91.88 & 92.80 & 77.27 & 63.33 & & 90.01 \\ 

    FedFR (N=2) & \textbf{91.55} & \textbf{91.57} & \textbf{76.97} & \textbf{49.40} & & \textbf{93.26} & & 94.25 & 96.00 & 85.60 & 70.93 & &  \textbf{94.99} \\ 
    
    \hline
    \vspace{-1cm}

    \end{tabular}
    \end{center}
\end{table*}



\section{Experiments}
\label{sec:experiments}


In this section, we first define a new benchmark for FedFR. Then we present the benchmark results and ablation studies.




\subsection{FedFR Benchmark}
\label{sec:benchmark}

We construct a new benchmark for federated unsupervised domain adaptation for face recognition, termed as FedFRBench. Table \ref{tab:evaluation-data} presents the statistics of training and testing datasets of the benchmark. 

\textbf{Training} \; FedFRBench contains around 87K labeled Caucasian identities in the source domain and 7K unlabeled African identities in the target domain. It is representative of the real-world scenario that the source domain has a large amount of labeled data while the target domain has limited unlabeled data. We construct source domain data of the benchmark from the MS-Celeb-1M dataset (MS1M) \cite{guo2016ms1m}. MS1M is a large-scale dataset, containing 100K identities, 82\% of which are Caucasian. Due to space constraints, we provide more details on construction in the supplementary.


\textbf{Evaluation} \; We measure the verification and identification accuracy of models using data from Racial Faces in-the-Wild (RFW) dataset \cite{wang2019rfw} in Table \ref{tab:evaluation-data}. We use around 3K Caucasian and 3K African identities to evaluate the performance of the source and target domains, respectively. For face verification, we present verification accuracy and true acceptance rate (TAR) at false acceptance rates (FARs) of 0.1, 0.01, and 0.001. For face identification, we report the rank-1 accuracy by matching a query to a gallery of images. 

\subsection{Implementation Details}

We implement FedFR and conduct experiments using PyTorch. For network architecture, we choose ResNet-34 as the backbone and the ArcFace \cite{deng2019arcface} as the loss. All experiments are conducted in clusters of eight NVIDIA V100 GPUs. We simulate FL by training each client in a GPU and use the PyTorch GPU communication backend to simulate the server aggregation. We assume that target domain clients have an equal amount of data; $K$ clients are simulated by splitting 7K African identities evenly to $K$ partitions. For example, when simulating 4 target domain clients, each client would contain 1,750 unlabeled data of African identities.


\subsection{Benchmark Results}

We compare FedFR with models obtained by other methods:
(1) \textit{Source-Only}: the baseline model obtained from supervised training with the source domain data. 
(2) \textit{Target-Only}: the model obtained from supervised training with the target domain data.
(3) \textit{Merge}: the model obtained from supervised training with data merged from both domains, presenting possible upper bound without data sharing constraint.
(4) \textit{Fine-tune}: the model fine-tuned on \textit{Source-Only} using pseudo labels predicted from clustering. 
(5) \textit{DAN} \cite{long2015dan}: the model trained by domain adaptation network with maximum mean discrepancies (MMD) loss, which is the classic discrepancy-based domain adaptation method.
(6) \textit{DANN} \cite{ganin2016dann}: the model trained by domain adversarial neural network, which is the classic adversarial-based domain adaptation method.

Table \ref{tab:fedfr-l} reports the experiment results on verification accuracy, verification rate at FAR=0.1, 0.01, and 0.001, and identification at rank-1. It shows that the benchmark is representative for the domain shift problem in the real-world scenario: 11\% gap on rank-1 accuracy between source and target domains (82.13\% vs 93.52\%) when trained with \textit{Source-Only} (only Caucasian data). Despite that DAN and DANN preserve the performance on the Caucasian dataset, they hardly improve the performance on the African dataset. On the contrary, FedFR outperforms all other models on the target domain (African), meanwhile maintaining comparable performance on the source domain. It even outperforms the possible upper bound \textit{Merge} model, demonstrating its effectiveness and significance. We run experiments with $d = 0.9$ and $\lambda = 0.01$ on the simplest federated setting with $N=2$ clients and a more complex setting with $N=5$ clients.


\subsection{Ablation Studies}

We investigate the effects of various components and the enhanced clustering algorithm in FedFR with ablation studies. 

\textbf{Ablation on FedFR} \; We compare the following four methods, each with an additional component based on the previous one: 
(1) \textit{Pre-training (P)}: supervised training on the source domain data. (2) \textit{P + Clustering (C)}: fine-tuning the pre-trained model using the target domain data with pseudo labels predicted from clustering. (3) \textit{P + C + FL}: conducting federated learning based on the pre-trained model and clustering. (4) \textit{P + C + FL + DCL}: conducting federated learning similar to (3) while with domain constraint loss (DCL).

Table \ref{tab:ablation} shows that the performance on the target domain (African) improves with each added component and reaches the peak with all components. Our proposed FL approach significantly boosts the performance by over 10\% and DCL further elevates it by around 4\% (at FAR-0.001). These results demonstrate the effectiveness and significance of FedFR.

\begin{figure}[t]
  \begin{center}
  \includegraphics[width=\linewidth]{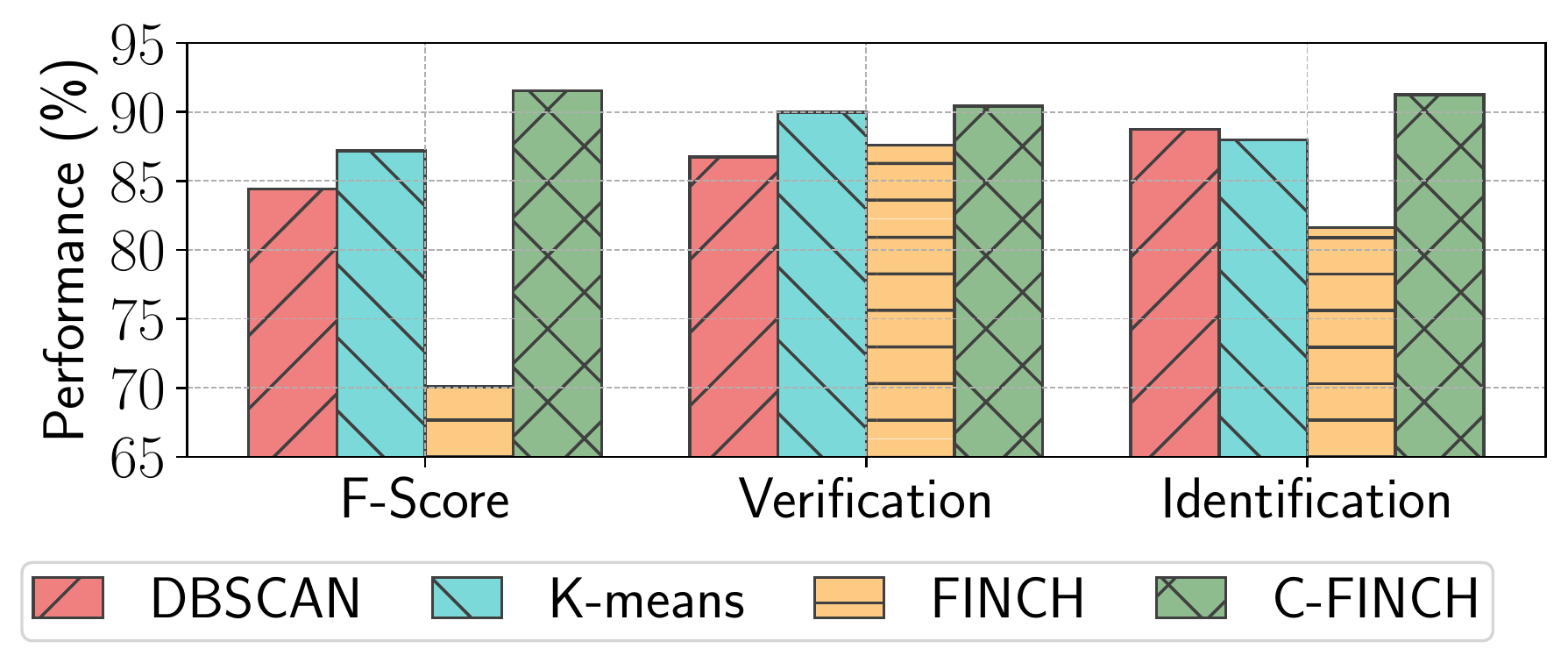}
  \end{center}
  \vspace{-0.5cm}
      \caption{Performance comparison of different clustering algorithms. Our C-FINCH outperforms other clustering methods even without relying on prior knowledge of datasets.}
  \label{fig:clustering}
  \vspace{-0.2cm}
\end{figure}

\textbf{Comparison of Clustering Algorithms} \; Figure \ref{fig:clustering} illustrates that our C-FINCH achieves the best performance (F-Score), leading to better verification and identification accuracy of FedFR on the target domain. Compared with FINCH \cite{sarfraz2019finch} (without the distance constrain), C-FINCH is superior by over 10\%. Compared with K-means \cite{macqueen1967k-means} and DBSCAN \cite{ester1996dbscan}, C-FINCH performs better and does not rely on prior knowledge about the cluster size, which is hard to obtain in reality.


\begingroup
\setlength{\tabcolsep}{0.3em}
\begin{table}[t]
\caption{Ablation study of FedFR. Each added component improves the performance on the benchmark.}
\label{tab:ablation}
\begin{center}
\begin{tabular}{lcccccc}
\hline
\multicolumn{1}{c}{\multirow{2}{*}{Training Method}} &
\multicolumn{5}{c}{African (\%)}
\\
\cline{2-6}
\multicolumn{1}{c}{} &
\multicolumn{1}{c}{Acc} &
\multicolumn{1}{c}{0.1} &
\multicolumn{1}{c}{0.01} &
\multicolumn{1}{c}{0.001} &
\multicolumn{1}{c}{Rank-1}
\\
\hline
Pre-training (P) & 86.65 & 83.27 & 55.93 & 37.80 & 82.13 \\ 
P + Clustering (C) & 88.72 & 87.10 & 62.23 & 35.95 & 84.05 \\ 
P + C + FL & 91.12 & 91.10 & 70.73 & 45.5 & 92.47 \\ 
P + C + FL + DCL & \textbf{91.55} & \textbf{91.57} & \textbf{76.97} & \textbf{49.40} & \textbf{93.26} \\ 
\hline
\vspace{-1cm}
\end{tabular}
\end{center}
\end{table}
\endgroup

\section{Conclusion}

In this paper, we present a novel federated unsupervised domain adaptation approach for face recognition, \textit{FedFR}, to address the domain shift problem under privacy constraints. FedFR jointly optimizes the distance-constrained hierarchical clustering algorithm and domain-constrained federated learning to elevate performance. Extensive experiments on the newly constructed benchmark demonstrate the effectiveness and significance of FedFR. Since the source domain normally contains more data than the target domain in reality, we propose FedFR based on this scenario. In the future, we consider evaluating other face datasets and exploring another scenario that the source domain contains less data than the target domain. Uneven data amounts in target domain clients will also be taken into consideration.

\vspace{1em}

\noindent\textbf{Acknowledgements} This study is in part supported by the RIE2020 Industry Alignment Fund – Industry Collaboration Projects (IAF-ICP) Funding Initiative, as well as cash and in-kind contribution from the industry partner(s); Singapore MOE under its Tier 1 grant call, Reference number RG96/20; Nanyang Technological University, Reference Number NTU–ACE2020-01.

\bibliographystyle{IEEEbib}
\bibliography{icme2022template}

\end{document}